# M2EF-NNs: Multimodal Multi-instance Evidence Fusion Neural Networks for Cancer Survival Prediction

Hui Luo[a,b], Jiashuang Huang[c,*], Hengrong Ju[c], Tianyi Zhou[c], Weiping Ding [a,c,*]

[a] Faculty of Data Science, City University of Macau, Macao 999078

[b] School of Information and Management, Guangxi Medical University, Nanning, Guangxi 530021

[c] School of Artificial Intelligence and Computer Science, Nantong University, Nantong, Jiangsu 226019

**Abstract**

Accurate cancer survival prediction is crucial for assisting clinical doctors in formulating treatment plans. Multimodal data, including histopathological images and genomic data, offer complementary and comprehensive information that can greatly enhance the accuracy of this task. However, the current methods, despite yielding promising results, suffer from two notable limitations: they do not effectively utilize global context and disregard modal uncertainty. In this study, we put forward a neural network model called M2EF-NNs, which leverages multimodal and multi-instance evidence fusion techniques for accurate cancer survival prediction. Specifically, to capture global information in the images, we use a pre-trained Vision Transformer (ViT) model to obtain patch feature embeddings of histopathological images. Then, we introduce a multimodal attention module that uses genomic embeddings as queries and learns the co-attention mapping between genomic and histopathological images to achieve an early interaction fusion of multimodal information and better capture their correlations. Subsequently, we are the first to apply the Dempster-Shafer evidence theory (DST) to cancer survival prediction. We parameterize the distribution of class probabilities using the processed multimodal features and introduce subjective logic to estimate the uncertainty associated with different modalities. By combining with the Dempster-Shafer theory, we can dynamically adjust the weights of class probabilities after multimodal fusion to achieve trusted survival prediction. Finally, Experimental validation on the TCGA datasets confirms the significant improvements achieved by our proposed method in cancer survival prediction and enhances the reliability of the model.

## 1. Introduction

Due to factors such as population aging, environmental pollution, unhealthy lifestyle habits, and changes in dietary patterns, cancer has become a significant global public health issue (Siegel et al., 2021). Histopathological images are widely regarded as the gold standard for cancer prognosis due to their pivotal role in guiding treatment decisions and

*Corresponding author

Email addresses: luohui@gxmu.edu.cn(Hui Luo), hjshdym@ntu.edu.cn(Jiashuang Huang),
juhengrong@ntu.edu.cn(Hengrong Ju), choutyear@stmail.ntu.edu.cn(Tianyi Zhou), ding.wp@ntu.edu.cn(Weiping Ding)

forecasting patient outcomes (Li et al., 2021; Liu et al., 2023; Shao et al., 2020; Shao et al., 2021; Wang et al., 2019). Survival prediction is an essential aspect of cancer prognosis research. However, relying solely on histopathological changes cannot objectively and accurately reflect the occurrence and development of the disease due to individual variations and cancer heterogeneity. With the advancements in molecular biology, there is now substantial research evidence indicating that many molecular pathological changes occur in the early stages of cancer, even before apparent morphological changes in cancer tissue. Moreover, even among cancer patients with the same histopathological type, there can be completely different molecular alterations within the cancer, leading to variations in drug treatment response and prognosis. Consequently, cancer experts often combine histopathology information with genomics information to predict patient survival outcomes (Chen et al., 2020; Gallego, 2015; Li et al., 2022; Wu et al., 2023). However, it is important to note that histopathological images and genomic data are heterogeneous, originating from different sources. Therefore, it is crucial to consider and address the heterogeneity among various modalities, as well as the unique information inherent in each modality. Fortunately, the emergence of Whole Slide Imaging (WSI) and breakthroughs in deep learning methods have provided exciting possibilities for the effective integration of histopathology and genomic information. This integration allows for better quantification of the tumor microenvironment and enables improved prediction of prognosis.

Recently, researchers have introduced a variety of deep learning-based methods aimed at enhancing the performance of cancer survival prediction through the integration of data from diverse modalities. For instance, Mobadersany et al. proposed GSCNN, an innovative method that effectively addresses the challenges posed by tumor heterogeneity in WSIs. GSCNN utilizes convolutional neural networks (CNNs) and incorporates image sampling and risk filtering techniques to mitigate the noise in WSIs resulting from tumor heterogeneity. By integrating genomic data and histopathological images, GSCNN achieves promising predictive performance in cancer survival prediction (Mobadersany et al., 2018). Cheerla et al. proposed a cancer survival prediction method that integrates pathological images and genomic data. They designed dedicated networks to extract features from different modalities and fused the multimodal features using a fully connected network, combined with a loss function based on the Cox proportional hazards model for model training. The experiments demonstrated that this method effectively predicts the prognosis of various cancers, particularly for cancers with limited training data, showing significant performance improvements (Cheerla & Gevaert, 2019). Hao et al. introduced PAGE-Net, a deep learning model that improves upon the path-based sparse deep neural network Cox-PASNet by incorporating genome-specific layers. Experimental results validated the superior predictive performance of this method compared to using only histopathological images and Cox-PASNet (Hao et al., 2019). Chen et al. presented a novel pathology fusion method for integrating histopathology and genomic data. The method utilized CNNs and graph convolutional networks (GCNs) to train morphological features from histopathological images and self-normalizing neural networks (SNNs) to train genomic features. They then simplified the output representation with a gate-based attention mechanism to reduce noise. Finally, they employed the Kronecker product to combine the deep features from histopathology and genomics, enabling the prediction of patient survival outcomes. The experiments yielded promising results (Chen et al., 2020). Wang et al. proposed the GPDBN, a fusion method that aims to enhance cancer survival prediction performance by taking into account the relationships within and between different modalities (Wang et al., 2021). Additionally, Lv et al. proposed a method called TransSurv for colorectal cancer survival analysis. This model effectively integrates the intra-modal and inter-modal features of multi-modal data, including genomic data, histopathology images, and clinical information (Lv et al., 2022). Subsequently, Chen et al. put forward a novel architecture named MCAT, which combines multiple instance learning and the Transformer model, to investigate intra-modality interactions. This model employs attention mechanisms to learn the interaction between images and genomics, thereby visually demonstrating the interpretability of multi-modality interactions (Chen et al., 2021). Furthermore, Xie et al. proposed a method called GC-SPLeM for predicting gastric cancer survival by learning from multi-modal data. This method incorporates a multi-modal attention module to aggregate information from different modalities and utilizes a GNN model to capture associations among patients, thereby mitigating the impact of missing data (Xie et al., 2022).

The current methods, despite yielding promising results, suffer from two notable limitations: they do not effectively utilize global context and disregard modal uncertainty. Specifically, traditional methods preserve limited global information and utilize patch features obtained from pre-trained ResNet models on ImageNet. In comparison to ViT models, CNN models lack spatial positional information and retain less global context. These structural differences result in variations in out-of-distribution generalization capabilities (Raghu et al., 2021; Zhang et al., 2022). However, survival prediction is a challenging ordinal regression task aimed at predicting the relative risk of cancer mortality, requiring complex interactions across instances and between instances throughout the entire WSI. Also, traditional approaches primarily focus on improving accuracy by leveraging the complementarity of different modalities while neglecting uncertainty, which may lead to unreliable predictions. The quality of data often varies across samples due to the presence of varying amounts of information and potential noise in histopathological images and genomic data (Argelaguet, 2021; Inoue & Yagi, 2020). Existing methods either simply treat each modality as equally important or adjust the weights of different modalities to fixed values, integrating them into a shared representation for subsequent tasks. Although these methods are effective, they often overlook uncertainty and are insufficient to capture the dynamic noise in multimodal data, potentially resulting in unreliable predictions. Therefore, ensuring the reliability of multimodal integration and final decision-making is also crucial.

To address these challenges, we propose Multimodal Multi-instance Evidence Fusion Neural Networks (M2EF-NNs) for cancer survival prediction. Our proposed method is capable of capturing comprehensive global information while considering the uncertainties associated with individual modalities. Specifically, we employ a pre-trained ViT model on ImageNet to extract feature embeddings from histopathology images. Then, using genomic embeddings as queries, we learn the co-attention mapping between the genomic features and histopathological images to achieve early interaction and fusion of multimodal information. Subsequently, we integrate the multimodal information at the evidence level using Dempster-Shafer theory and dynamically adjust the weights of class probability distribution after multimodal fusion to achieve trusted survival prediction. The primary contributions of this study are summarized as follows:

1) To effectively capture the global information in the histopathological images, we employ a pre-trained ViT model trained on the ImageNet dataset. Specifically, we sliced the histopathological images at a magnification scale of 20x and extracted features using the pre-trained model.

2) We introduce a multimodal attention module to fuse the processed histopathology images and genomic data. By learning the co-attention mapping between the genomic features and histopathology images, we achieve early interaction and fusion of multimodal information, enabling us to better capture their correlations.

3) We consider the uncertainty of each modality in the cancer survival prediction by using the DST. We parameterize the distribution of class probabilities using the processed multimodal features and introduce subjective logic to estimate the uncertainty associated with different modalities. By combining with the Dempster-Shafer theory, we can dynamically adjust the weights of class probabilities after multimodal fusion to achieve trusted survival prediction. Based on the information available to us, our work is the first to explore the application of the Dempster-Shafer theory in cancer survival prediction.

4) Experimental validation on the TCGA dataset confirms the significant improvements achieved by our proposed method in cancer survival prediction and enhances the reliability of the model.

The remaining sections of this paper are organized as follows. the related works are introduced in Section 2. Section 3 presents a detailed description of the M2EF-NNs model. Section 4 reports the experimental results and comparisons. Section 5 presents a discussion, as well as limitations and future works. Section 6 serves as the conclusion of the paper.

## 2. Related Work

*2.1. Multiple Instance Learning (MIL)*

In pathology, the heterogeneity and high resolution nature of WSI makes it challenging to directly label the entire image and use it for training. Moreover, only a small portion of the WSI is usually relevant to the disease. Therefore, many methods have started to adopt weakly supervised learning approaches to deal with pathology images. Currently, these weakly supervised methods based on MIL and other set-based deep learning techniques have achieved remarkable progress in the field of WSI applications. Researchers such as Edwards, Storkey, and Zaheer were among the pioneers in proposing neural network architectures based on set data structures (Edwards & Storkey, 2016; Zaheer et al., 2017). Subsequently, Ilse et al. further extended the set-based architecture by introducing attention mechanisms and applying them to WSIs (Ilse et al., 2018). To further enhance predictive performance, Yao et al. proposed an attention-guided deep MIL network, which combines multiple instance learning and attention mechanisms for survival analysis in cancer patients, offering good interpretability (Yao et al., 2020). Recently, Chen et al. introduced MCAT, a novel structure that combines multiple instance learning and Transformer models for early fusion interaction of modalities, further exploring the interactions within modalities and achieving good predictive performance (Chen et al., 2021). In summary, while MIL-based pathology analysis methods may not fully capture the complex interactions between instances, they have been able to effectively address the needle-in-a-haystack problem in pathology. These methods provide powerful tools for pathology researchers to tackle the challenges of whole-slide digital pathology images and offer more accurate disease analysis and prediction.

*2.2. Vision Transformer (ViT)*

Google introduced the Transformer architecture in 2017 as a deep learning framework. Initially designed to tackle machine translation tasks, it utilizes a self-attention mechanism and incorporates a multi-head attention mechanism (Vaswani et al., 2017). In 2020, Dosovitskiy et al. introduced the architecture to image classification for the first time, proposing the ViT (Dosovitskiy et al., 2020). This work demonstrated that image classification is not necessarily dependent on traditional CNNs and becoming a benchmark method for many computer vision tasks. Subsequent research has made improvements to the ViT, including enhancements to feature embeddings, image patch segmentation, and other modules, further advancing the evolution of Transformers in the realm of computer vision. Recent studies on ViT have primarily focused on robustness (Bhojanapalli et al., 2021), the impact of self-attention on the models (Caron et al., 2021), model improvements (Liu et al., 2021), and comparisons with CNN models (Raghu et al., 2021; Zhang et al., 2022). In the medical domain, histopathology images exhibit complex composition, where some abnormal images consist mostly of abnormal patches, while others contain only a small fraction of abnormal patches. Therefore, models used for histopathology image feature extraction tasks must possess strong capabilities to extract global information. The ViT architecture introduces positional encoding, improving accessibility to global information and exhibiting excellent out-of-distribution generalization and prominent feature representation capabilities. As a result, ViT has demonstrated outstanding performance in histopathology images segmentation and classification tasks (Chen et al., 2022; Gao et al., 2021; Huang et al., 2023; Luo et al., 2023; Wang et al., 2023). For example, Gao et al. introduced i-ViT, a method designed to learn robust representations of histopathological images for tasks involving subtyping papillary renal cell carcinoma. This approach focuses on extracting detailed features from patches to enhance the quality of the learned representations (Gao et al., 2021). Chen et al. presented GasHisTransformer, a model that combines the strengths of ViT and CNN architectures to achieve automatic global detection of gastric cancer images (Chen et al., 2022). Huang et al. proposed a novel CSF Transformer model, designed to efficiently integrate patch embeddings from diverse fields of view and learn cross-scale contextual relationships (Huang et al., 2023). Although Transformers have made significant progress in histopathology image segmentation and classification, their application in survival analysis is still under development and requires further exploration.

*2.3. Dempster-Shafer Evidence Theory (DST)*

DST, initially proposed by Dempster, is a theory that deals with belief functions (Dempster, 1967). It serves as an extension of Bayesian theory by incorporating subjective probabilities (Dempster, 1968). It has been developed as a general framework for modeling uncertainty in cognition (Shafer, 1976). Unlike Bayesian neural networks that indirectly model uncertainty through parameter sampling, DST directly models uncertainty. It enables the combination of beliefs from multiple sources using different fusion operators, resulting in new beliefs that incorporate all available evidence (Jøsang & Hankin, 2012). In data classification, the belief functions and Dempster's combination rule of evidence theory have been combined with various classification algorithms to handle uncertainty (Denoeux, 2000; Peñafiel et al., 2020; Sensoy et al., 2018; Zhang et al., 2023; Zhu et al., 2021). Denoeux et al. proposed a multi-layer neural network that combines the Dempster rule for adaptive pattern classification (Denoeux, 2000). Sensoy et al. applied evidence theory to quantify uncertainty in deep convolutional neural network classification and constructed an evidence-deep neural network (Sensoy et al., 2018). Furthermore, evidence-based deep neural networks have been used to build algorithms for uncertain data classification (Gao et al., 2022; Ghesu et al., 2019; Tardy et al., 2019). Recently, Han et al. introduced a trusted multimodal classification method that combines the DS evidence theory and subjective logic. It parameterizes the distribution parameters of class probabilities as Dirichlet distributions using evidence from different views and estimates the uncertainty of different viewpoints (Han et al., 2022). In order to ensure the reliability of multimodal integration and the final prediction of histopathology and genomics, we are the first to introduce the Dempster-Shafer evidence theory for trusted multimodal survival prediction.

## 3. Method

In this section, we will introduce our overall framework, as shown in Fig. 1. The M2EF-NNs proposed in this paper contain three main modules, including the multimodal multiple instance feature extraction (refer to Section 3.1 for details), the multimodal feature fusion (refer to Section 3.2 for details), and the DST-based trusted survival prediction (refer to Section 3.3 for details).

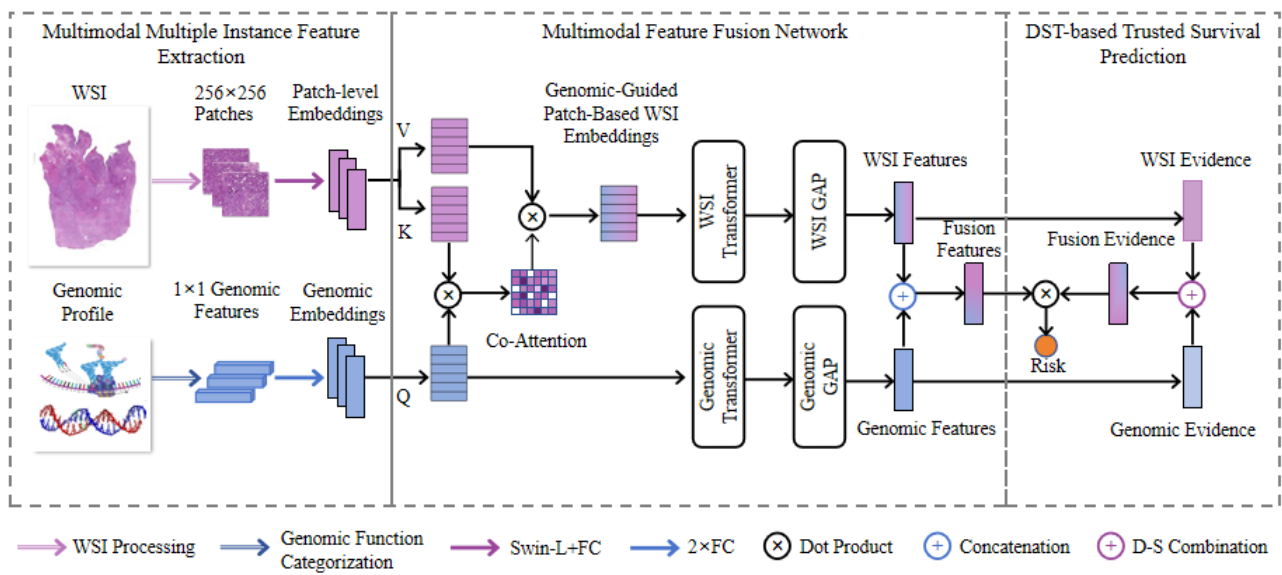

Fig. 1. Overview of M2EF-NNs. It consists of three parts: multimodal multiple instance feature extraction, multimodal feature fusion, and DST-based Trusted Survival Prediction. 1) Multimodal multiple instance feature extraction: The part is to extract features from multimodal

data. Specifically, patches are sampled from WSI with 20x magnification and then fed them into a pre-trained Swin-L model to extract patch features at the instance level. Second, based on the biological function, genomic profiles are classified, and instance-level genomic features are extracted. 2) Multimodal feature fusion: In this part, multiple strategies are employed to fuse features from different modalities. First, we learn the co-attention mapping between the genomic features and histopathology images to achieve early interaction and fusion of multimodal information. Second, multi-instance features are aggregated using global attention pooling, transforming instance-level features into image-level features. Finally, the features from different modalities are fused using a concatenation method to obtain the final fused feature vector. 3) DST-based Trusted Survival Prediction: In this stage, the Dempster-Shafer theory is utilized to integrate multimodal information at the evidence level and dynamically adjust the weights of fused class probabilities after multimodal fusion for trusted survival prediction.

### 3.1. Multimodal Multiple Instance Feature Extraction

In multiple instance learning, training samples are organized in the form of instance bags, where each bag consists of multiple instances and is assigned a label indicating its class. In our study, we constructed corresponding instance bags for each patient's histopathology images and genomic data. Each instance bag contains a varying number of instances. The label for each instance bag represents the total survival time of the patient in months. Additionally, we used a label to indicate the survival status, where 0 represents death and 1 represents survival.

**Multiple Instance Feature Extraction for Histopathological:** To handle large-scale WSIs, we adopted a two-stage MIL approach for feature extraction. First, we needed to crop the complete WSI into multiple equal-sized patches and extract features from each patch to obtain instance-level features for subsequent tasks. Considering the high resolution of WSIs, processing the entire image at once in memory is challenging. Therefore, we opted for a staged approach. Before model training, we first used a pre-trained model for feature extraction, saving the extracted features for subsequent model training. The entire feature extraction process is illustrated in Fig. 2. We selected WSIs at a magnification scale of 20x for analysis. First, for each WSI, we performed tissue segmentation employing the method described in (Lu et al., 2021). The segmented tissue regions were then cropped into 256×256 patches, ensuring no spatial overlap between the patches. For each patch, we extracted a 1536-dimensional feature embedding $h \in \mathbb{R}^{1536 \times 1}$ using the pre-trained Swin-L model (Liu et al., 2021) trained on ImageNet. Subsequently, the feature embedding was mapped to a 256-dimensional space through a fully connected layer (FC layer). Finally, we packaged each patch-level embedding into an instance bag $H_{bag} \in \mathbb{R}^{M \times 256}$, where $M$ is the instance size of the bag and the instance size of each WSI is not unique.

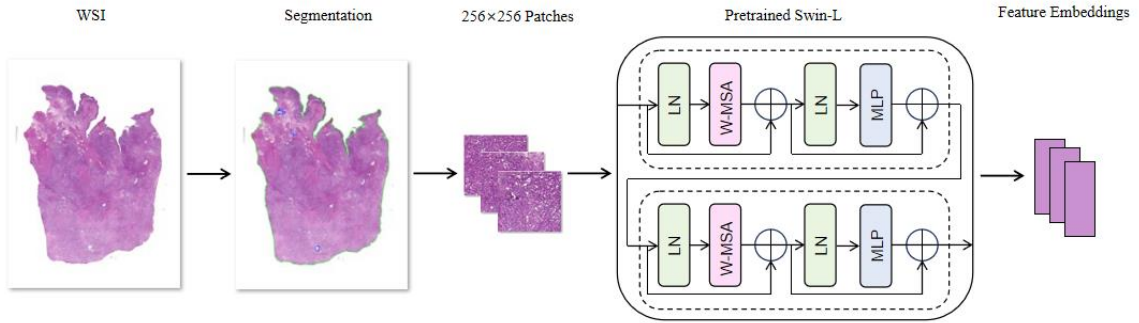

Fig. 2. The specific framework of WSI feature extraction

**Multiple Instance Feature Extraction for Genomics:** Genomic data primarily consists of gene mutation status, copy number variations, and gene expression data, which are typically represented as 1×1 attributes. In our study, we utilized the semantic information within genes and grouped genes with similar biological functions into one instance for feature extraction. We divided the gene sets using six functional category labels obtained from (Liberzon et al., 2015; Subramanian et al., 2005). These sets are tumor suppressor, tumor occurrence, protein kinase, cell differentiation, transcription, and cytokine and growth. The size of each gene set may vary. To process these genomic data, we employed two fully connected layers (FC layers). The first FC layer was used to partition all gene attributes and assign them to their corresponding gene sets. The second FC layer mapped each gene set to a 256-dimensional feature embedding $\{G_n \in \mathbb{R}^{256 \times 1}\}_{n=1}^{N}$. These feature embeddings were then packaged into a genomic bag $G_{bag} \in \mathbb{R}^{N \times 256}$. Here, $N$ was set to 6, indicating that each genomic bag contained six instances corresponding to six functional category label sets.

*3.2. Multimodal Feature Fusion Network*

In multimodal feature fusion, our goal is to integrate features from different modalities to achieve a more comprehensive and robust feature representation than a single modality alone. In this stage, our primary emphasis is on the multimodal fusion of histopathology images and genomic features. The objective is to construct a unified feature representation for subsequent survival prediction. This process involves three steps: cross-modal early fusion, multiple instance aggregation, and multimodal late fusion.

**Cross-modal early fusion:** First, inspired by the work of (Chen et al., 2021) and (Vaswani et al., 2017), we found that we can use genomic embeddings $G_{bag} \in \mathbb{R}^{N \times 256}$ as a substitute for the query Q in the Transformer attention mechanism. We then multiply the query with the histopathology image embeddings $H_{bag} \in \mathbb{R}^{M \times 256}$ to obtain a co-attention matrix $A_{coat}$ between the two modalities, achieving early interaction fusion. Here, our goal is to map histopathology image embeddings $H_{bag}$ to a set of genomic-guided histopathology image embeddings $\widehat{H}_{bag}$. This mapping can be represented as:

$$CoAt_{G \rightarrow H}(G, H) = softmax\left(\frac{QK^\intercal}{\sqrt{d_k}}\right) = softmax\left(\frac{W_q G H^\intercal W_k^\intercal}{\sqrt{256}}\right) W_v H \rightarrow A_{coat} W_v H \rightarrow \widehat{H}. \tag{1}$$

Where $W_q$, $W_k$, and $W_v$ are trainable weight matrices that are multiplied with the query $G_{bag}$ and key-value pairs $(H_{bag}, H_{bag})$, respectively, and $A_{coat} \in \mathbb{R}^{N \times M}$ is the co-attention matrix used to compute the weighted average of $H_{bag}$. By employing this approach, we achieve an early fusion of histopathology image features and genomic features, allowing them to influence and interact with each other, thereby extracting richer multimodal features. Furthermore, in the experiments, the batch size of $Q$ is 6, while the batch sizes of $K$ and $V$ are generally larger than $Q$. As a result, aggregating $\widehat{H}_{bag} \in \mathbb{R}^{N \times 256}$ is computationally more efficient than aggregating $H_{bag} \in \mathbb{R}^{M \times 256}$, significantly reducing the computational cost. The obtained $\widehat{H}_{bag}$ is essentially a WSI feature embedding that is enhanced with the attention of genomic information.

**Multiple instance aggregation:** Then, we pass the obtained genomic-guided histopathology image embeddings $\widehat{H}_{bag}$ and the genomic feature matrix $G_{bag}$ through two Transformer encoder layers and a global attention-based pooling layer for multiple instance aggregation. In the instance aggregation, we first compute their attention scores as follows:

$$A_{ih} = \frac{\exp\{W_{\rho h}[\tanh(V_{\rho h} h_i^\intercal) \odot sigmod(U_{\rho h} h_i^\intercal)]\}}{\sum_{ih=1}^{N} \exp\{W_{\rho h}[\tanh(V_{\rho h} h_i^\intercal) \odot sigmod(U_{\rho h} h_i^\intercal)]\}}, \tag{2}$$

$$A_{ig} = \frac{\exp\{W_{\rho g}[\tanh(V_{\rho g} g_i^\intercal) \odot sigmod(U_{\rho h} g_i^\intercal)]\}}{\sum_{ig=1}^{N} \exp\{W_{\rho g}[\tanh(V_{\rho g} g_i^\intercal) \odot sigmod(U_{\rho h} g_i^\intercal)]\}}, \tag{3}$$

where $W_{\rho h}$, $V_{\rho h}$, $U_{\rho h}$, $W_{\rho g}$, $V_{\rho g}$, and $U_{\rho g}$ are all trainable parameters. Then, we concatenate all the instance vectors $h$ and $g$ within the bags $\widehat{H}_{bag}$ and $G_{bag}$ to form the feature vector as shown below:

$$R_h = ReLU\left(W_{\zeta h} \sum_{ih=1}^{N} A_{ih} h_i\right), \tag{4}$$

$$R_g = ReLU\left(W_{\zeta g} \sum_{ig=1}^{N} A_{ig} g_i\right), \tag{5}$$

where $W_{\zeta h}$ and $W_{\zeta g}$ are also trainable parameters.

**Multimodal late fusion:** Finally, we pass the aggregated genomic-guided histopathology image feature vector and the genomic feature vector to an MLP (Multi-Layer Perceptron) for late fusion. This MLP comprises two layers with ReLU activation functions. Ultimately, we obtain a 256-dimensional feature vector as follows:

$$R_{fusion} = MLP(R_h \oplus R_g), \tag{6}$$

where $R_{fusion}$ includes both histopathological images and genomic information.

*3.3. DST-based Trusted Survival Prediction*

In the survival prediction task, we adopted the same model as (Chen et al., 2021) and modeled the survival time using discrete time intervals. We divided the survival time into four non-overlapping intervals based on its quantiles. Therefore, the discrete survival time corresponds to four labeled categories, then K is equal to 4. Next, we input the unified feature representation constructed through late fusion into a fully connected layer (FC layer). In this FC layer, for each time interval, we can calculate a survival risk score that represents the patient's survival risk within that interval. We use the sigmoid activation function to fit the patient's survival risk score as follows:

$$s_{risk} = sigmod(FC(R_{fusion})), \tag{7}$$

where $s_{risk}$ is a four-dimensional vector, representing the survival risk for each category.

Inspired by the research of (Han et al., 2022), to ensure the reliability of multimodal integration and the final prediction of histopathology and genomics, we used evidence from different modes to parameterize the distribution of category probability into Dirichlet distribution and introduced subjective logic to estimate the uncertainty of different modes, combined with Dempster-Shafer theory, the weights of category probabilities after multimodal fusion are dynamically adjusted to make trusted multimodal survival prediction. Specifically, firstly, we can get non-negative evidence vectors $e_h$ and $e_g$ from the feature embeddings aggregated by histopathology and genomics through a full connection layer and a Softplus activation layer respectively as follows:

$$e_h = Softplus(FC(R_h)), e_g = Softplus(FC(R_g)). \tag{8}$$

Subjective logic links the evidence with the parameters of Dirichlet distribution, for $\alpha = e + 1$. Accordingly, the parameters of Dirichlet distribution of the two modes $\alpha_h$ and $\alpha_g$ can be obtained as follows:

$$\alpha_h = e_h + 1, \alpha_g = e_g + 1. \tag{9}$$

$S_h$ and $S_g$ are Dirichlet strengths which are calculated as follows:

$$S_h = \sum_{i=1}^{K}(e_{kh} + 1) = \sum_{i=1}^{K} \alpha_h, S_g = \sum_{i=1}^{K}(e_{kg} + 1) = \sum_{i=1}^{K} \alpha_g. \tag{10}$$

Subsequently, we calculate the confidence mass and overall uncertainty for each modality separately as follows:

$$b_{kh} = \frac{e_{kh}}{S_h} = \frac{\alpha_{kh} - 1}{S_h}, \quad b_{kg} = \frac{e_{kg}}{S_g} = \frac{\alpha_{kg} - 1}{S_g}, \tag{11}$$

$$u_h = \frac{K}{S_h}, \quad u_g = \frac{K}{S_g}, \tag{12}$$

where $u_h, u_g, b_h, b_g \geq 0$, and meet $u_h + \sum_{k=1}^{K} b_{kh} = 1$, $u_g + \sum_{k=1}^{K} b_{kg} = 1$. After obtaining the evidence and uncertainties from individual modalities, we utilized the DST to combine evidence from various sources and generate a degree of belief that incorporates all available evidence. The specific combination rules followed the approach outlined by (Han et al., 2022). Through the utilization of the Dempster-Shafer combination rule, we can derive the final confidence quality for each category and assess the overall uncertainty as follows:

$$b_k = \frac{1}{1-C}(b_{kh}b_{kg} + b_{kh}u_g + b_{kg}u_h), u = \frac{1}{1-C}u_h u_g. \tag{13}$$

Thereby, we can estimate the fused evidence from multiple modalities and determine the corresponding parameters of the Dirichlet distribution as follows:

$$S = \frac{K}{u}, e_k = b_k \times S \text{ and } \alpha_k = e_k + 1. \tag{14}$$

Then, the final evidence for each category, obtained through evidence fusion, is normalized using the sigmoid function. This normalization process allows us to obtain the weight parameters for the risk scores of each category, derived from the multimodal fusion network. The calculation of the final risk scores is as follows:

$$o_{risk} = sigmod(e) \cdot s_{risk} . \tag{15}$$

In the final risk score $o_{risk}$, the uncertainty of each modality is taken into account. The sizes of the final evidence for each class, obtained through evidence fusion, are used to dynamically adjust the risk scores obtained from the modality fusion network. The greater the evidence accumulated, the higher the weight assigned, leading to an increased final risk score for that particular class. Conversely, the less evidence there is, the smaller the weight assigned to the corresponding risk score, leading to a lower final risk score for the class. Compared to other methods, our approach reduces decision-making risks and improves the reliability of the model. When updating the parameters of the survival model, we followed the approach of (Chen et al., 2021) by considering the log-likelihood (Zadeh & Schmid, 2020) that takes into account the patient's survival status.

## 4. Experiments and Results

*4.1. Dataset descriptions*

To validate our proposed method, we collected data on Bladder Urothelial Carcinoma (BLCA) and Glioma (GBMLGG) from The Cancer Genome Atlas (TCGA). We matched the diagnostic WSIs of each patient with their genomic information and clinical information, including survival time labels and review status. The statistical information for the two constructed datasets is shown in Table 1. For the BLCA dataset, we had a total of 373 patients, 437 WSIs, and 3392 genes. For the GBMLGG dataset, which includes both Glioblastoma Multiforme and Brain Lower Grade Glioma, we had 567 patients, 1014 WSIs, and 2723 genes.

Table 1. Data statistics of datasets

| Dataset | BLCA | GBMLGG |
|---|---|---|
| Patient Number | 373 | 567 |
| WSI Number | 437 | 1014 |
| Gene Number | 3392 | 2723 |

*4.2. Implementation details*

We trained our model using an NVIDIA GTX 4090 GPU. Firstly, we sampled slices at a magnification scale of 20x and extracted 1536-dimensional image feature embeddings using the Swin-L model pre-trained on ImageNet. Then, we applied multiple fully connected layers to transform the genomic data into genomic feature embeddings. Next, we fed the extracted histopathology feature embeddings and patient genomic feature embeddings into the multimodal fusion network. We first learned a co-attention mapping between the genomic data and WSIs. Then, we processed the WSI and genomic features separately using two visual transformers and an attention pooling layer. The preprocessed features were subsequently combined through concatenation and fed into multiple fully connected layers with sigmoid activations to obtain the concatenated risk scores. Simultaneously, the evidence fusion of different modalities was performed using the final evidence from the modal fusion network as the weight parameters for the risk scores of each class obtained from the modal fusion network. During the training phase, we employed the Adam optimizer with a learning rate of $2 \times 10^{-4}$ and weight decay of $1 \times 10^{-5}$.

*4.3. Evaluation metrics*

In our experiments, we utilized the concordance index (c-Index) as an evaluation metric. The concordance index, ranging from 0 to 1, measures the predictive performance of the model, where a higher value indicates better performance and vice versa. The concordance index quantifies the correlation between risk scores and survival times and is a commonly used evaluation metric for survival prediction. The calculation formula is as follows:

$$c - Index = \frac{1}{n} \sum_{i \in \{1 \cdots N | \sigma_i = 1\}} \sum_{s_j > s_i} I[X_i \hat{\beta} > X_j \hat{\beta}]. \tag{16}$$

Where $n$ is the number of comparable pairs, $I[\cdot]$ is the indicator function, and $s$ are the observed values.

In addition to the concordance index, We also employed the log-rank test (Bland & Altman, 2004) to evaluate the statistical significance of differences between the two survival curves. The log-rank test allows for the comparison of survival curve differences between different groups, such as the predicted high-risk and low-risk groups. This test helps determine whether the risk stratification predicted by the model has statistical significance. Furthermore, we employed the Kaplan-Meier estimation and the predicted risk distribution to visualize patient stratification (Kaplan & Meier, 1958). The Kaplan-Meier estimation is a non-parametric approach utilized to estimate survival curves based on information regarding a patient's survival time and events, such as death or recurrence. By visualizing the predicted risk distribution and survival curves, we can intuitively demonstrate how the model stratifies patients.

*4.4. Methods for Comparison*

In comparisons with state-of-the-art methods on two cancer datasets, we first compare the proposed M2EF-NNs method with a single-modality genomic-based approach and three single-modality MIL methods based on WSIs, including 1) self-normalizing neural network (SNN) method, 2) sum pooling-based (Deep Sets) method, 3) global attention pooling-based (Attention MIL) method, and 4) cluster-based (DeepAttnMISL) method. Subsequently, we further compare M2EF-NNs with four state-of-the-art multimodal MIL methods, i.e., 1) sum pooling-based (Deep Sets) method, 2) global attention pooling-based (Attention MIL) method, 3) cluster-based (DeepAttnMISL) method, and 4) co-attention-based (MCAT) method. Now we briefly summarize these competing methods as follows.

1) SNN (Klambauer et al., 2017): SNN is a self-normalizing neural network architecture used for genomic single-line comparison. It utilizes self-normalizing activation functions and weight initialization methods to enhance the training effectiveness and robustness of the network.

2) Deep Sets (Zaheer et al., 2017): Deep Sets is one of the earliest set-based neural network architecture. It is a method that performs feature aggregation and pooling at the instance level. This approach can handle input sets with an uncertain number of instances and models the set using a pooling operation with a variable order.

3) Attention MIL (Ilse et al., 2018): Attention MIL is a multiple instance learning architecture based on attention mechanisms. It extends the concept of set-based learning and uses global attention pooling instead of sum pooling to aggregate instances in a deep set. This method can adaptively learn the importance weights of different instances to better capture key instances within the set.

4) DeepAttnMISL (Yao et al., 2020): DeepAttnMISL is a survival analysis method based on multiple-instance learning and attention mechanisms. It first clusters instance features into K clusters using the K-Means method and then extracts features from each cluster using convolutional neural networks. Finally, the cluster instance features are aggregated using global attention pooling. This method demonstrates good performance and interpretability in explaining survival-related pathology images.

5) MCAT (Chen et al., 2021): MCAT is an advanced interpretable and weakly supervised multimodal survival prediction architecture. It is based on multi-instance learning and the Transformer model, to explore intra-modality interactions. This model learns the interaction between images and genes using attention mechanisms and visually demonstrates the interpretability of multi-modality interactions.

There are two major strategies in M2EF-NNs, i.e., 1) using a pre-trained Swin-L model to extract pathological image features, and 2) performing trusted survival prediction based on the DST. To investigate the effectiveness of these strategies, we further compare M2EF-NNs with its three variants, including 1) uses the pre-trained Swin-L model for extracting pathological image features but does not employ the survival predictor based on the D-S evidence theory (denoted as MCAT (Swin-L)), 2) does not use the pre-trained Swin-L model for extracting pathological image features but employs the survival predictor based on the D-S evidence theory (denoted as M2EF-NNs (Resnet 50)), and 3) uses the pre-trained Swin-L model for extracting pathological image features and employs the survival predictor based on the D-S evidence theory (denoted as M2EF-NNs (Swin-L)). Additionally, for the purpose of multimodal comparison with M2EF-NNs, we only employed concatenation fusion as a late fusion mechanism to integrate histopathological and genomic features. For all methods, we used the same 5-fold cross-validation splits, training hyperparameters, and loss functions on the two cancer datasets.

*4.5. Experimental Results*

Table 2 shows the experimental results on two cancer datasets using different methods. From the table, we can find that M2EF-NNs outperform DeepSets, Attention MIL, and DeepAttnMISL in terms of single-modality data training methods. The c-Index performance of M2EF-NNs improved by 18.52%, 12.20%, and 13.93% in all benchmark tests, respectively. This indicates that M2EF-NNs are more accurate in survival prediction tasks compared to these methods. Relative to the genomic baseline, M2EF-NNs achieved a performance improvement of 5.90%. This suggests that M2EF-NNs exhibit better survival prediction capability on genomic data. M2EF-NNs show improvements in all benchmark tests compared to their corresponding single-modality tasks, which is consistent with similar work that utilizes multimodal fusion to enhance supervised learning tasks. This further validates the effectiveness of multimodal data fusion in improving survival prediction performance. Compared to the multimodal methods of DeepSets, Attention MIL, DeepAttnMISL, and MCAT, M2EF-NNs achieved an overall c-Index improvement of 5.29%, 5.00%, 3.95%, and 2.08%, respectively. This indicates that M2EF-NNs have improved overall performance in multimodal survival prediction tasks.

Table 2. The results of comparison experimental by different methods using c-Index values

| Methods | Model | BLCA | GBMLGG | Overall |
|---|---|---|---|---|
| Unimodal | SNN (Genomic Only) | 0.583±0.029 | 0.807±0.014 | 0.695 |
|  | DeepSets (WSI Only) | 0.497±0.013 | 0.744±0.041 | 0.621 |
|  | Attention MIL (WSI Only) | 0.541±0.075 | 0.771±0.032 | 0.656 |
|  | DeepAttnMISL (WSI Only) | 0.537±0.024 | 0.754±0.025 | 0.646 |
| Multimodal | DeepSets | 0.597±0.036 | 0.801±0.046 | 0.699 |
|  | Attention MIL | 0.601±0.024 | 0.801±0.028 | 0.701 |
|  | DeepAttnMISL | 0.606±0.025 | 0.809±0.014 | 0.708 |
|  | MCAT | 0.624±0.023 | 0.817±0.024 | 0.721 |
|  | M2EF-NNs | **0.651±0.021** | **0.821±0.034** | **0.736** |

Furthermore, we utilized the 50th percentile as the risk index in the Kaplan-Meier survival analysis to categorize patients into low-risk and high-risk groups. Fig. 3 presents the results of the analysis. The Kaplan-Meier survival curves clearly demonstrate that the M2EF-NNs method outperforms other multimodal methods in both datasets, indicating its superior performance. In the BLCA dataset, the log-rank test p-values for DeepSets, Attention MIL, DeepAttnMISL, MCAT, and M2EF-NNs were 1.904e-03, 1.053e-02, 2.939e-03, 3.045e-04, and 2.130e-06 respectively. Among them, M2EF-NNs showed the most significant improvement. In the GBMLGG dataset, the log-rank test p-values for DeepSets, Attention MIL, DeepAttnMISL, MCAT, and M2EF-NNs were 4.586e-25, 1.731e-23, 1.120e-24, 3.789e-29, and 3.590e-31 respectively. Once again, M2EF-NNs demonstrated the most significant improvement. Therefore, these experiments indicate that the proposed M2EF-NNs method can enhance the performance of cancer survival prediction.

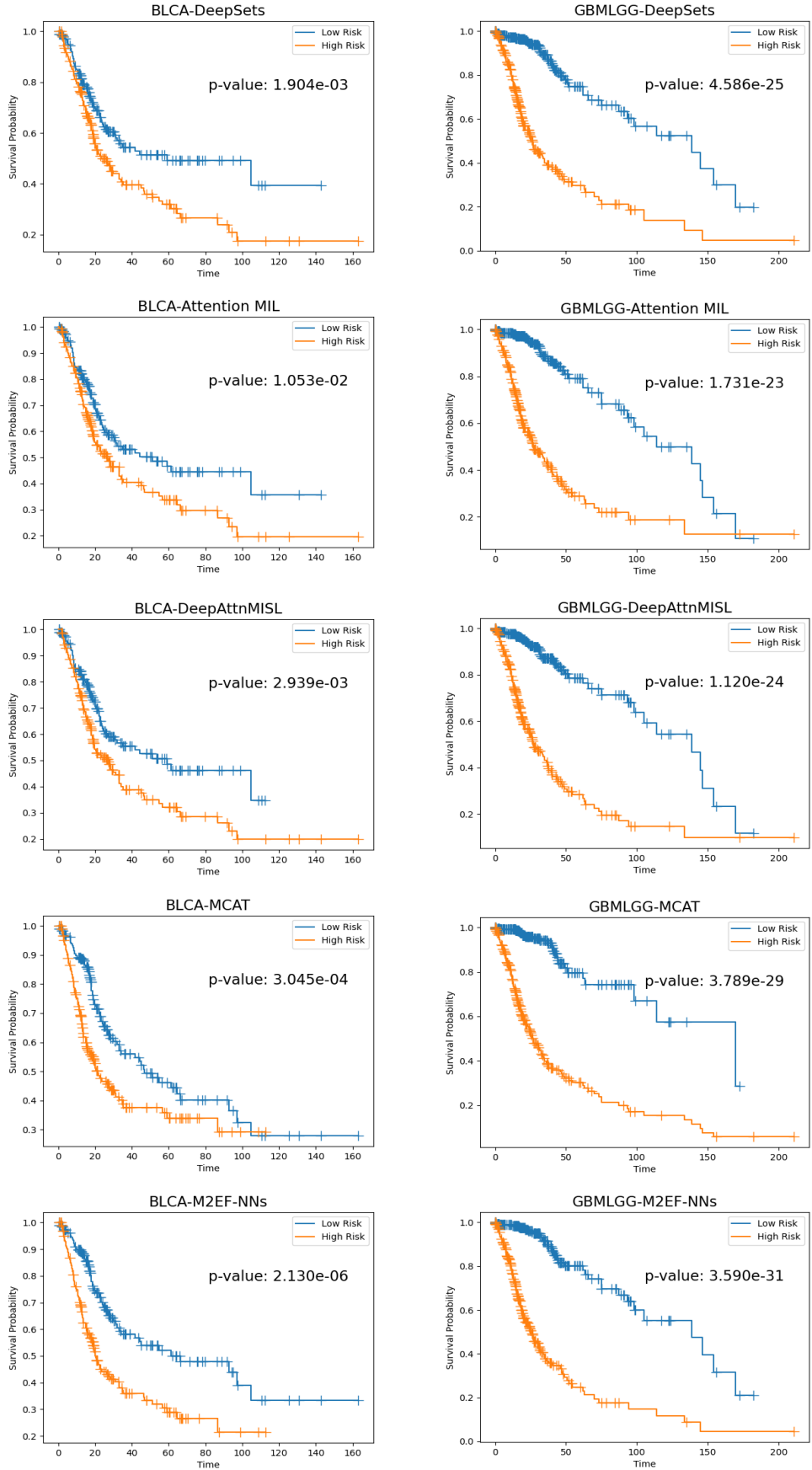

Fig. 3. Kaplan-Meier survival curves of the multimodal methods.

*4.6. Ablation Studies*

To further evaluate the effectiveness of the pre-trained Swin-L model for patch feature extraction and the trusted survival predictor based on the DST, we conducted ablation experiments. Table 3 presents the results of the ablation experiments. The results demonstrate that both using the Swin-L model for patch feature extraction and the multimodal evidence fusion-based trusted survival predictor contributes to the overall c-index. Specifically, comparing M2EF-NNs (Swin-L) with M2EF-NNs (ResNet 50), we found that the model using the Swin-L model for patch feature extraction outperformed the method using the ResNet 50 model, with an average c-index of 0.736. These findings demonstrate that the Swin-L model effectively captures representative features from medical images. Furthermore, the performance difference between M2EF-NNs (Swin-L) and MCAT (Swin-L) validates the effectiveness of the trusted survival predictor based on the DST.

Table 3. The results of ablation studies using c-Index values

| Methods | BLCA | GBMLGG | Overall |
| --- | --- | --- | --- |
| MCAT (Swin-L) | 0.633±0.030 | 0.817±0.030 | 0.725 |
| M2EF-NNs (Resnet 50) | 0.641±0.016 | 0.818±0.022 | 0.730 |
| M2EF-NNs (Swin-L) | **0.651±0.021** | **0.821±0.034** | **0.736** |

Furthermore, upon examining the Kaplan-Meier survival curves in Fig. 4 and the corresponding log-rank test p-values, it becomes evident that our proposed M2EF-NNs method, which leverages the pre-trained Swin-L model for patch feature extraction and the trusted survival predictor based on the DST, significantly improves its performance. In the BLCA dataset, M2EF-NNs (Swin-L) exhibits the most noteworthy log-rank test p-value of 2.130e-06 compared to M2EF-NNs (ResNet 50) and MCAT (Swin-L). Similarly, in the GBMLGG dataset, M2EF-NNs (Swin-L) demonstrates the most remarkable log-rank test p-value of 3.590e-31 compared to M2EF-NNs (ResNet 50) and MCAT (Swin-L).

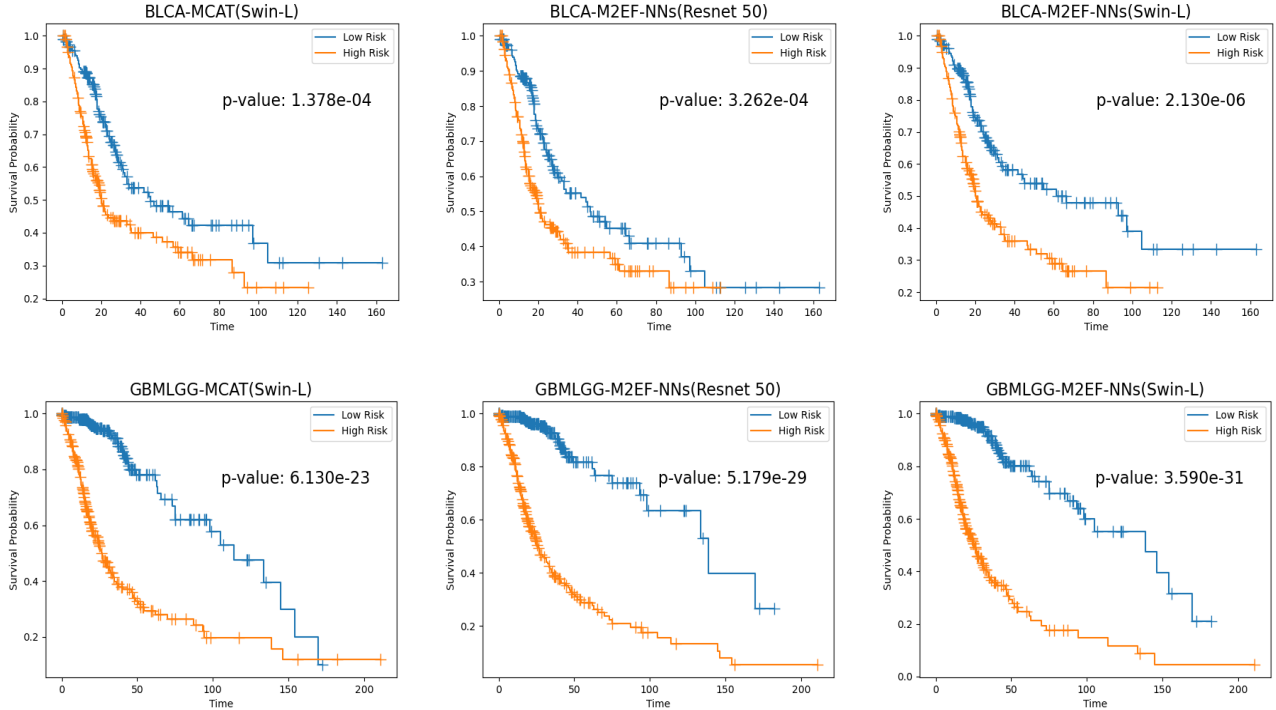

Fig. 4. Kaplan–Meier survival curves of ablation study

*4.7. Multimodal Interpretability*

In addition to the improved c-Index and significant patient stratification, our M2EF-NNs method is highly interpretable. We utilize genomic embeddings to guide the computation of shared attention weights for histopathological patches. These shared attention weights are overlaid onto the corresponding spatial locations in the original whole-slide images (WSI) to construct genomic-guided visual concept-attention heatmaps. As shown in Fig. 5, in both low-risk and high-risk cases of TCGA-BLCA, the genomic-guided heatmaps reflect several known genotype-phenotype relationships in cancer pathology. Specifically, in BLCA, the genomic-guided visual concept-attention heatmaps often reflect normal stroma, glands, and adipocytes, which are associated with tumor suppression, protein kinases, and cellular differentiation. Across all cases, the cellular differentiation embeddings predominantly focus on tumor-associated stroma, while tumor suppression and protein kinase embeddings primarily concentrate on adipocyte-adjacent stroma and glandular structures. In the tumor and transcriptomic embeddings, regions with high attention weights are localized to aggressive, higher-grade tumor morphologies, such as dense tumor cells and tumor-infiltrated stroma. In the cytokine embeddings, high attention regions are centered around immune cells and tumor cells infiltrating the normal stroma.

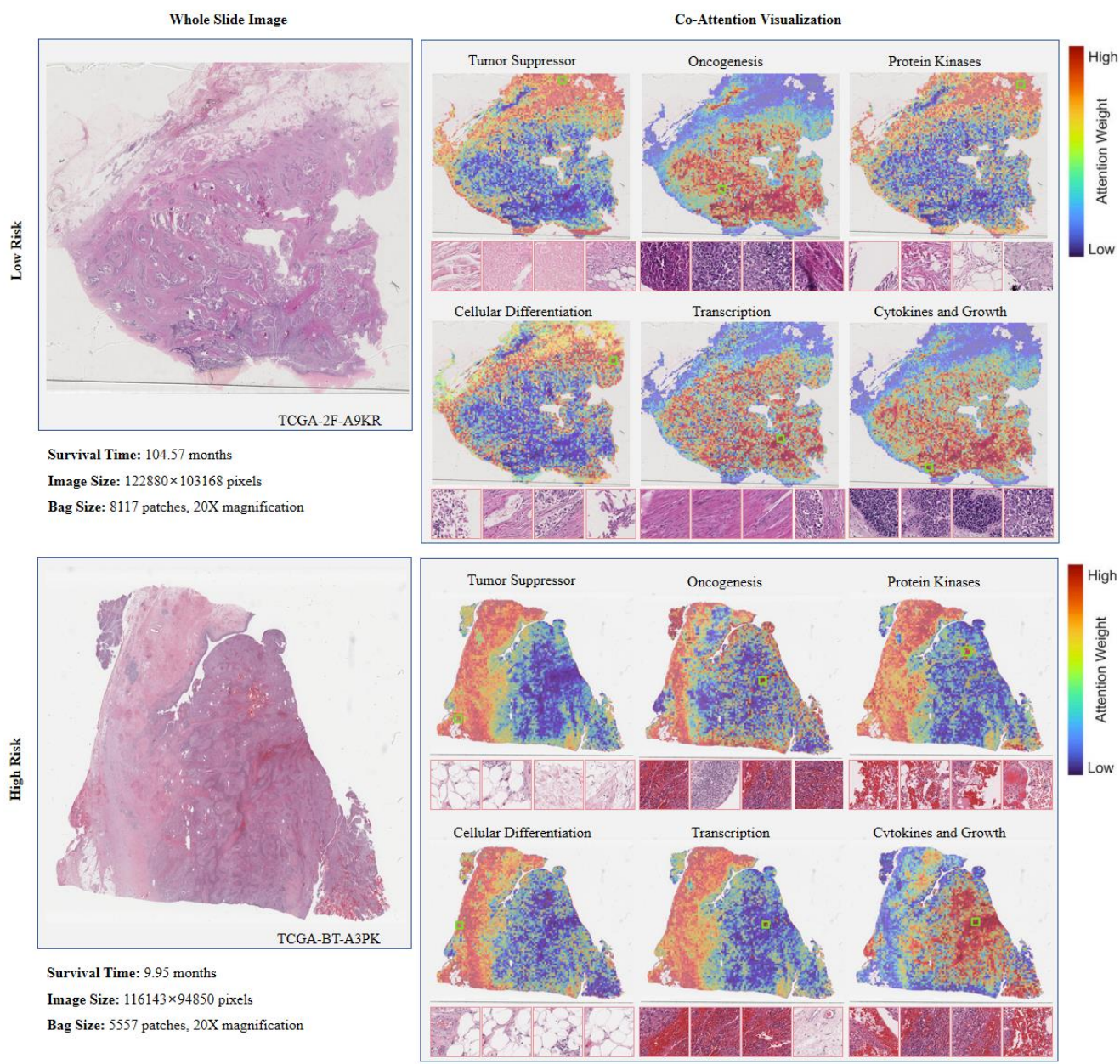

Fig. 5. Attention heatmaps with corresponding high attention patches

These observations indicate that through genomics-guided shared attention heatmaps, the M2EF-NNs method is able to capture some known genotype-phenotype relationships in cancer pathology, aiding in the understanding of cancer development and pathological features. This highly interpretable approach can provide medical researchers and clinicians with deeper insights to support cancer diagnosis, treatment, and prognosis assessment.

## 5. Discussion

In the diagnosis and prognosis assessment of cancer, it is typically necessary to obtain multiple biomarkers to accurately evaluate the disease status and progression stage. Among these biomarkers, histopathological images and genomic data are two primary types. While there have been some studies proposing multimodal survival prediction methods that integrate histopathology images and genomics data, achieving good results, there are two significant limitations: they cannot effectively utilize global context and they disregard the uncertainty of modalities. To extract effective predictive features from these multimodal data for multimodal integration and ensure the reliability of the final decision, we propose Multimodal Multi-instance Evidence Fusion Neural Networks (M2EF-NNs) for cancer survival prediction. Our proposed method is capable of capturing comprehensive global information while considering the uncertainties associated with individual modalities. Specifically, we employ a pre-trained Swin-L model to extract features from histopathology images. Then, using genomic embeddings as queries, we learn the co-attention mapping between the genomic features and histopathology images to achieve early interaction and fusion of multimodal information. Subsequently, we integrate the multimodal information at the evidence level using Dempster-Shafer theory and dynamically adjust the weights of the class probability distribution after multimodal fusion to achieve trusted survival prediction.

In line with similar studies that employ multimodal fusion to enhance supervised learning tasks, M2EF-NNs demonstrates improvements over its respective single-modal counterparts in all benchmark evaluations. This further validates the effectiveness of multimodal data fusion in improving survival prediction performance. Compared to the multimodal methods of DeepSets, Attention MIL, DeepAttnMISL, and MCAT, the M2EF-NNs method exhibits overall performance improvement for multimodal survival prediction. This provides a new computational tool for cancer survival prediction. Additionally, ablation experiments further validate the effectiveness of the pre-trained Swin-L model in capturing representative patch features and the trusted survival predictor based on the D-S evidence theory.

Our study has two limitations. Firstly, it only analyzes pathological images at a single scale to reduce training time costs. WSIs are typically organized in a pyramid structure, with data scanned at different magnification levels stored in different pyramid levels. Pathologists continuously review and analyze pathological images at different magnifications while moving across the microscope or terminal screen, rather than focusing on local or fixed magnifications. Analyzing slices of pathological images at a single scale may not adequately represent the overall heterogeneous microenvironment of the tumor. Secondly, the separation of the WSI feature extraction process from the final prediction stage in M2EF-NNs does not allow for end-to-end training. In future research, it would consider incorporating multi-scale feature fusion and employing more powerful approaches for patch filtering in WSI, optimizing WSI feature extraction in conjunction with the multimodal fusion network.

## 6. Conclusion

In this work, we propose Multimodal Multi-instance Evidence Fusion Neural Networks (M2EF-NNs) for cancer survival prediction. Our proposed method is capable of capturing comprehensive global information while considering the uncertainties associated with individual modalities. Specifically, we employ a pre-trained Swin-L model to extract features from histopathology images. Then, using genomic embeddings as queries, we learn the co-attention mapping between the genomic features and histopathology images to achieve early interaction and fusion of multimodal information. Subsequently, we integrate the multimodal information at the evidence level using Dempster-Shafer theory and dynamically adjust the weights of the class probability distribution after multimodal fusion to achieve trusted survival prediction. The experimental results unequivocally demonstrate a substantial enhancement in cancer survival prediction performance with the proposed method, thereby significantly boosting the model's reliability.

# Acknowledgments

This work is supported in part by the National Natural Science Foundation of China (61976120, 62006128, 62102199), the Natural Science Foundation of Jiangsu Province (BK20231337), the Natural Science Key Foundation of Jiangsu Education Department (21KJA510004), the General Program of the Natural Science Foundation of Jiangsu Province Higher Education Institutions (20KJB520009), the Basic Science Research Program of Nantong Science and Technology Bureau (JC2020141, JC2021122), the Postgraduate Research & Practice Innovation Program of Jiangsu Province ( SJCX21_1446, SJCX22_1615), and the Youth Science Foundation of Guangxi Medical University (GXMUYSF202432).